\documentclass[letter,12pt]{article}

\usepackage[margin=2.5cm]{geometry}

\usepackage{graphics}
\usepackage{graphicx} 
\usepackage{rotating}
\usepackage{dcolumn}
\usepackage{longtable}
\usepackage{multirow}
\usepackage{amsmath}
\usepackage{amssymb}
\usepackage{xspace}

\begin{document}

\title{The Outer Product Structure of Neural Network Derivatives}
\author{Craig Bakker, Michael J. Henry \& Nathan O. Hodas}
\maketitle

\begin{abstract}
In this paper, we show that feedforward and recurrent neural networks exhibit an outer product derivative structure but that convolutional neural networks do not.  This structure makes it possible to use higher-order information without needing approximations or infeasibly large amounts of memory, and it may also provide insights into the geometry of neural network optima.  The ability to easily access these derivatives also suggests a new, geometric approach to regularization.  We then discuss how this structure could be used to improve training methods, increase network robustness and generalizability, and inform network compression methods.
\end{abstract}

\section{Introduction}
\label{Introduction}

Gradient-based optimization is a foundational part of Deep Learning (DL) in that it makes it possible to train neural networks in an efficient manner.  Training then requires calculating the gradients of the loss function with respect to the weights at each layer of the network.  How these gradients are calculated, therefore, is very important.  Today, this is done using backpropagation \cite{rumelhart86jsr}, which is a version of the chain rule from calculus.  Backpropagation is highly efficient, but it is typically only used to calculate first-order derivatives; higher-order derivatives are generally considered too expensive to calculate \cite{goodfellow16bk}.  Essentially, backpropagation exploits the directed structure of the network to make the calculation of these gradients easier.  This begs the question, of course, of whether there is more structure that could be exploited to improve algorithm performance or reduce computational cost.

Such structure does exist for feedforward networks.  As we will show in Section \ref{The Outer Product Structure of Deep Network Derivatives}, these derivatives have, on a per-sample basis, an outer product structure to them.  We will demonstrate this for first-order derivatives and show how it extends to higher-order derivatives.  Because of this outer product structure, the widely held assumption that the Hessian is too large to use may not actually hold.

We are not the first to note or be interested in these derivative properties.  In describing the backpropagation algorithm, Rumelhart et al. \cite{rumelhart86jsr} use index notation (see Section \ref{The Outer Product Structure of Deep Network Derivatives}) in writing out their equations.  The index notation clearly shows the outer product structure for first-order derivatives, but the authors do not comment on this.  Mizutani and Dreyfus \cite{mizutani08jsr} and Naumov \cite{naumov17jsr} note this structure, while Xie et al. \cite{xie15jsr} and Zhang et al. \cite{zhang17jsr} explicitly take advantage of it to speed up their gradient computations in what they refer to as Sufficient Factor Broadcasting.

This structure extends to feedforward network Hessians.  Surprisingly, though, this is not noted in most of the papers that look at calculating the Hessian or approximations thereto (e.g., \cite{bishop92jsr,buntine94jsr}).  Naumov \cite{naumov17jsr} does note this, but the derivations presented there are incorrect by being incomplete.  The Hessian calculations there only consider `block-diagonal' elements of the Hessian -- i.e., Hessian components for weights in the same layer -- and not cross-derivatives between weights from different layers.  Those terms add some complexity to the Hessian structure, as will be apparent in our calculations.  Unfortunately, this omission means that Naumov's eigenvalue calculations also do not hold, though they could be considered as a kind of block-diagonal approximation.  Naumov also focuses on a sum-of-squared error loss function.  We will provide a more general and complete analysis here.

\section{The Outer Product Structure of Deep Network Derivatives}
\label{The Outer Product Structure of Deep Network Derivatives}

Second-order derivatives are not widely used in DL, and where they \emph{are} used, they are typically estimated.  These derivatives can be calculated analytically, but this is not often done because of scalability constraints.  If we write out the first and second derivatives, though, we can see that they have an outer product structure to them.  When a matrix has low rank (or less than full rank), it means that the information contained in that matrix (or the operations performed by that matrix) can be fully represented without needing to know every entry of that matrix.  An outer product structure is a special case of this, where an $m$x$n$ matrix $\mathbf{A}$ can be fully represented by two vectors $\mathbf{A} = \mathbf{u} \mathbf{v}^T$.  We can then calculate, store, and use second-order derivatives exactly in an efficient manner by only dealing with the components needed to represent the full Hessians rather than dealing with those Hessians themselves.  Doing this involves some extra calculations, but the storage costs are comparable to those of gradient calculations.

In this section, we will illustrate this structure for a feedforward network, of arbitrary depth and layer widths, consisting of ReLUs in the hidden layers.  A feedforward network with arbitrary activation functions has somewhat more complicated derivative formulae, but those derivatives still exhibit an outer product structure.  That structure also does not depend on the form of the objective function or whether a softmax is used, and it is present for recurrent layers as well.  The complete derivations for these cases are given in Appendix \ref{Low-Rank Derivations for Deep Networks}.

In our calculations, we make extensive use of index notation with the summation convention \cite{ivancevic07bk}.  In index notation, a scalar has no indices ($v$), a vector has one index ($\mathbf{v}$ as $v^i$ or $v_i$), a matrix has two ($\mathbf{V}$ as $V^{ij}$, $V^i_j$, or $V_{ij}$), and so on.  The summation convention holds that repeated indices in a given expression are summed over unless otherwise indicated.  For example, $\mathbf{a}^T \mathbf{b} = \sum_i a^i b_i = a^i b_i$.  The pair of indices being summed over will often consist of a superscript and a subscript; this is a bookkeeping technique used in differential geometry, but in this context, the subscripting or superscripting of indices will not indicate covariance or contravariance.  We have also adapted index notation slightly to suit the structure of deep networks better: indices placed in brackets (e.g. the $k$ in $v^{\left(k\right),j}$) are not summed over, even if repeated, unless explicitly indicated by a summation sign.  A tensor convention that we \emph{will} use, however, is the Kronecker delta: $\delta^{ij}$, $\delta^i_j$, or $\delta_{ij}$.  The Kronecker delta is the identity matrix represented in index notation: it is 1 for $i=j$ and 0 otherwise.  The summation convention can sometimes be employed to simplify expressions containing Kronecker deltas.  For example, $\delta^j_i v^i = v^j$ and $\delta_{ij} V_{jk} = V_{ik}$.  

\subsection{Feedforward Network Derivative Calculations}
\label{Feedforward Network Derivative Calculations}

Our example here is a generic feedforward network with ReLU activation functions in $n$ hidden layers, a softmax at the output layer, and categorical cross-entropy as the objective function.  The softmax and categorical cross-entropy are not important for our calculations, but they are commonly used and we include them here for concreteness' sake; we consider ReLUs because they are widely used and have some convenient properties.  Table \ref{Nomenclature -- Formulation} provides a nomenclature for our deep network definition, and Equations \ref{network start}-\ref{network end} define the network.

\begin{table}[htp]
\centering
\caption{Nomenclature -- Formulation}
\label{Nomenclature -- Formulation}
\vspace{6pt}
\begin{tabular}{c|c}
Quantity													& Description \\
\hline
$n$																& Number of hidden layers \\
$x^i$															& Vector of inputs for a single sample \\
$v^{\left(k\right),j}$						& Vector output of layer $k$ \\
$w^{\left(k\right),j}_i$					& Matrix of weights for layer $k$ \\
$\mathcal{A}\left(\cdot\right)$		& Activation function \\
$u^j_i$														& Matrix of output layer weights \\
$p^j$															& Vector of intermediate variables for the output layer \\
$\hat{y}^l$												& Vector of outputs for a single sample \\
$y^l$															& Vector of labels for a single sample \\
$f$																& Scalar objective function value for a single sample \\
$F$																& Scalar objective function
\end{tabular}
\end{table}

\begin{gather}
v^{\left(k\right),j} = \mathcal{A} \left( w^{\left(k\right),j}_i v^{\left(k-1\right),i}\right), \ k = 1, \ldots, n 
\label{network start}\\
\mathcal{A} \left(z\right) = \max \left(z,0\right) \\
 v^{\left(0\right),i} = x^i \\
p^j = u^j_i v^{\left(n\right),i} 
\label{p eqn} \\
\hat{y}^j = \frac{ \exp \left(p^j\right)}{\sum \limits_l \exp\left(p^l\right)} \\
f = - y^l \ln \hat{y}^l \\
F = E\left[f\right]
\label{network end}
\end{gather}

The relevant first derivatives for this deep network are

\begin{gather}
\mathcal{A}'\left(z\right) = \left\{ \begin{array}{cc}
1		&z>0 \\
0		&z<0 \end{array} \right. \\
\frac{\partial v^{\left(k\right),j}}{\partial w^{\left(l\right),s}_t} = \left\{ \begin{array}{cc}
0																																																	&l>k \\
\delta^j_s \mathcal{A}'\left(w^{\left(k\right),j}_i v^{\left(k-1\right),i}\right) v^{\left(k-1\right),t}		&l=k \\
\mathcal{A}'\left(w^{\left(k\right),j}_i v^{\left(k-1\right),i}\right) w^{\left(k\right),j}_q \frac{\partial v^{\left(k-1\right),q}}{\partial w^{\left(l\right),s}_t}		& l<k
\end{array} \right. 
\label{no j 1}
\end{gather}

\noindent where there is no summation over $j$ in Equation \ref{no j 1}.  We now define several intermediate quantities to simplify the derivation process:

\begin{gather}
\gamma^{\left(k\right),j}_s \equiv \delta^j_s \mathcal{A}'\left(w^{\left(k\right),j}_i v^{\left(k-1\right),i} \right) 
\label{no j 2}\\
\beta^{\left(k\right),j}_i \equiv \mathcal{A}'\left(w^{\left(k\right),j}_l v^{\left(k-1\right),l}\right) w^{\left(k\right),j}_i = \gamma^{\left(k\right),j}_s w^{\left(k\right),s}_i 
\label{no j 3}\\
\alpha^{\left(k,l\right),j_k}_{j_l} \equiv \left\{ \begin{array}{cc}
\prod \limits_{i=l+1}^k \beta^{\left(i\right),j_i}_{j_{i-1}} 	&k>l \\
\delta^i_j 																										&k=l\\
0																															&k<l
\end{array} \right. \\
\alpha^{\left(k,l\right),j_k}_{j_l} \alpha^{\left(l,m\right),j_l}_{j_m} = \alpha^{\left(k,m\right),j_k}_{j_m} \\
\eta^{\left(k,l\right),j}_i \equiv \alpha^{\left(k,l\right),j}_s \gamma^{\left(l\right),s}_i
\end{gather}

\noindent where there is no summation over $j$ in Equations \ref{no j 2} and \ref{no j 3}.  We can now complete our calculations of the first derivatives.

\begin{gather}
\frac{\partial v^{\left(k\right),j}}{\partial w^{\left(l\right),s}_t} = \left\{ \begin{array}{cc}
0																																																	&l>k \\
\eta^{\left(k,l\right),j}_s v^{\left(l-1\right),t}		&l\leq k 
\end{array} \right. \\
\frac{\partial p^j}{\partial u^l_k} = \delta^j_l v^{\left(n\right),k} \\
\frac{\partial p^j}{\partial v^{\left(n\right),i}} = u^j_i \\
\frac{\partial f}{\partial u^i_j} = \frac{\partial f}{\partial p^k} \frac{\partial p^k}{\partial u^i_j} = \frac{\partial f}{\partial p^i} v^{\left(n\right),j} 
\label{df/du}\\
\frac{\partial f}{\partial w^{\left(k\right),i}_j} = \frac{\partial f}{\partial p^l} \frac{\partial p^l}{\partial v^{\left(n\right),m}} \frac{\partial v^{\left(n\right),m}}{\partial w^{\left(k\right),i}_j} = \frac{\partial f}{\partial p^l} u^l_m \eta^{\left(n,k\right),m}_i v^{\left(k-1\right),j}
\label{df/dw} \\
\frac{\partial f}{\partial u^i_j} = \frac{\partial f}{\partial p^i} v^{\left(n\right),j} \\
\frac{\partial f}{\partial w^{\left(k\right),i}_j} = \frac{\partial f}{\partial p^l} u^l_m \eta^{\left(n,k\right),m}_i v^{\left(k-1\right),j}
\end{gather}

In calculating these expressions, we have deliberately left $\frac{\partial f}{\partial p^j}$ unevaluated.  This keeps the expression relatively simple, and programs like TensorFlow \cite{tensorflow} can easily calculate this for us.  Leaving it in this form also preserves the generality of the expression -- there is no outer product structure contained in $\frac{\partial f}{\partial p^j}$, and the outer product structure of the network as a whole is therefore shown to be independent of the objective function and whether or not a softmax is used.  In fact, as long as Equation \ref{p eqn} holds, any sufficiently smooth function of $p^j$ may be used in place of a softmax without disrupting the structure.  The one quantity that needs to be stored here is $\eta^{\left(n,k\right),j}_i$ for $k = 1, 2, \ldots, n-1$; it will be needed in the second derivative calculations.  Note, however, that this is roughly the same size as the gradient itself.  

We can now see the outer product structure: $\frac{\partial f}{\partial u^i_j}$ is the outer product (or tensor product) of the vectors $\frac{\partial f}{\partial p^i}$ and $v^{\left(n\right),j}$, and $\frac{\partial f}{\partial w^{\left(k\right),i}_j}$ is the outer product of $\frac{\partial f}{\partial p^l} u^l_m \eta^{\left(n,k\right),m}_i$ (which ends up being a rank-1 tensor) and $v^{\left(k-1\right),j}$.  The index notation makes the outer product structure clear.  Vectorizing the weights, as is often done \cite{goodfellow16bk} makes this structure much more difficult to see.  

We then start our second derivative calculations by considering some intermediate quantities:

\begin{gather}
\mathcal{A}''\left(z\right) = 0 \\
\frac{\partial \alpha^{\left(n,k\right),m}_q}{\partial w^{\left(r\right),s}_t} = \alpha^{\left(n,r\right),m}_a \mathcal{A}'\left(w^{\left(r\right),a}_p v^{\left(r-1\right),p}\right) \delta^a_s \delta^t_b \alpha^{\left(r-1,k\right),b}_q = \eta^{\left(n,r\right),m}_s \alpha^{\left(r-1,k\right),t}_q \\
\frac{\partial \gamma^{\left(k\right),j}_s}{\partial \left(\cdot\right)} = 0 \\
\frac{\partial \eta^{\left(n,k\right),m}_i}{\partial w^{\left(r\right),s}_t} = \frac{\partial \alpha^{\left(n,k\right),m}_q}{\partial w^{\left(r\right),s}_t} \gamma^{\left(k\right),q}_i \\
\frac{\partial^2 v^{\left(n\right),m}}{\partial w^{\left(k\right),i}_j \partial w^{\left(r\right),s}_t} = \left\{ \begin{array}{cc}
\frac{\partial \alpha^{\left(n,k\right),m}_q}{\partial w^{\left(r\right),s}_t} \gamma^{\left(k\right),q}_i v^{\left(k-1\right),j}	&r>k \\
0																																																																		& r=k \\
\eta^{\left(n,k\right),m}_i \frac{\partial v^{\left(k-1\right),j}}{\partial w^{\left(l\right),s}_t}																	&r<k 
\end{array} \right. \nonumber \\ \
= \left\{ \begin{array}{cc}
\eta^{\left(n,r\right),m}_s \eta^{\left(r-1,k\right),t}_i v^{\left(k-1\right),j}	&r>k \\
0																																									& r=k \\
\eta^{\left(n,k\right),m}_i \eta^{\left(k-1,r\right),j}_s v^{\left(r-1\right),t}	&r<k 
\end{array} \right.
\end{gather}

The second derivative of the ReLU vanishes, which simplifies the second derivative calculations significantly.  Technically, the second derivative is undefined at the origin, but the singularity is removable, and thus we can define the second derivative to be 0 at the origin.  We can then calculate the second-order objective function derivatives:

\begin{gather}
\frac{\partial^2 f}{\partial u^i_j \partial u^s_t} = \frac{\partial^2 f}{\partial p^k \partial p^l} \frac{\partial p^k}{\partial u^i_j} \frac{\partial p^l}{\partial u^s_t} = \frac{\partial^2 f}{\partial p^i \partial p^s} v^{\left(n\right),j} v^{\left(n\right),t} \\
\frac{\partial^2 f}{\partial u^i_j \partial w^{\left(k\right),s}_t} = \frac{\partial f}{\partial p^i} \frac{\partial v^{\left(n\right),j}}{\partial w^{\left(k\right),s}_t} + \frac{\partial^2 f}{\partial p^i \partial p^l} v^{\left(n\right),j} \frac{\partial p^l}{\partial v^{\left(n\right),m}} \frac{\partial v^{\left(n\right),m}}{\partial w^{\left(k\right),s}_t} \nonumber \\
= \frac{\partial f}{\partial p^i} \eta^{\left(n,k\right),j}_s v^{\left(k-1\right),t} + \frac{\partial^2 f}{\partial p^i \partial p^l} v^{\left(n\right),j} u^l_m \eta^{\left(n,k\right),m}_s v^{\left(k-1\right),t} \\
\frac{\partial^2 f}{\partial w^{\left(k\right),i}_j \partial w^{\left(r\right),s}_t} = \frac{\partial^2 f}{\partial p^l \partial p^q} \frac{\partial p^l}{\partial v^{\left(n\right),m}} \frac{\partial v^{\left(n\right),m}}{\partial w^{\left(k\right),i}_j} \frac{\partial p^q}{\partial v^{\left(n\right),a}} \frac{\partial v^{\left(n\right),a}}{\partial w^{\left(r\right),s}_t} + \frac{\partial f}{\partial p^l} \frac{\partial p^l}{\partial v^{\left(n\right),m}} \frac{\partial^2 v^{\left(n\right),m}}{\partial w^{\left(k\right),i}_j \partial w^{\left(r\right),s}_t} \nonumber \\
= \frac{\partial^2 f}{\partial p^l \partial p^q} u^l_m \eta^{\left(n,k\right),m}_i v^{\left(k-1\right),j} u^q_a \eta^{\left(n,r\right),a}_s v^{\left(r-1\right),t} \nonumber \\
+ \frac{\partial f}{\partial p^l} u^l_m \times \left\{ \begin{array}{cc}
\eta^{\left(n,r\right),m}_s \eta^{\left(r-1,k\right),t}_i v^{\left(k-1\right),j}	&r>k \\
0																																									& r=k \\
\eta^{\left(n,k\right),m}_i \eta^{\left(k-1,r\right),j}_s v^{\left(r-1\right),t}	&r<k 
\end{array} \right. \\
\frac{\partial^2 f}{\partial u^i_j \partial u^s_t} = \frac{\partial^2 f}{\partial p^i \partial p^s} v^{\left(n\right),j} v^{\left(n\right),t} \\
\frac{\partial^2 f}{\partial u^i_j \partial w^{\left(k\right),s}_t} = \frac{\partial f}{\partial p^i} \eta^{\left(n,k\right),j}_s v^{\left(k-1\right),t} + \frac{\partial^2 f}{\partial p^i \partial p^l} v^{\left(n\right),j} u^l_m \eta^{\left(n,k\right),m}_s v^{\left(k-1\right),t} \\
\frac{\partial^2 f}{\partial w^{\left(k\right),i}_j \partial w^{\left(r\right),s}_t} = \frac{\partial^2 f}{\partial p^l \partial p^q} u^l_m \eta^{\left(n,k\right),m}_i v^{\left(k-1\right),j} u^q_a \eta^{\left(n,r\right),a}_s v^{\left(r-1\right),t} \nonumber \\
+ \frac{\partial f}{\partial p^l} u^l_m \times \left\{ \begin{array}{cc}
\eta^{\left(n,r\right),m}_s \eta^{\left(r-1,k\right),t}_i v^{\left(k-1\right),j}	&r>k \\
0																																									& r=k \\
\eta^{\left(n,k\right),m}_i \eta^{\left(k-1,r\right),j}_s v^{\left(r-1\right),t}	&r<k 
\end{array} \right. \\
\end{gather}

Calculating all of the second derivatives requires the repeated use of $\frac{\partial^2 f}{\partial \mathbf{p}^2}$.  Evaluating that Hessian is straightforward given knowledge of the activation functions and objective used in the network, and storing it is also likely not an issue as long as the number of categories is small relative to the number of weights.  For example, consider a small network with 10 categories and 1000 weights.  In such a case, $\frac{\partial^2 f}{\partial \mathbf{p}^2}$ would only contain 100 entries -- the gradient would be 10 times larger.  We also have to store $\eta^{\left(n,k\right),i}_j$ values in order to calculate the derivatives.  In $\frac{\partial^2 f}{\partial \mathbf{w}^2}$, we also end up needing $\eta^{\left(r,k\right),i}_j$ for $r \neq n$.  In a network with $n$ hidden layers, we would then have $\frac{n \left(n-1\right)}{2}$ of the $\eta^{\left(r,k\right),i}_j$ matrices to store.  For $n=10$, this would be 45, for $n=20$, this would be 190, and so on.  

This aspect of the calculations does not seem to scale well, but in practice, it is relatively simple to work around.  It is still necessary to store $\eta^{\left(n,k\right),i}_j$, $k < n$, but $\eta^{\left(r,k\right),i}_j$, $r < n$, only actually shows up in one place, and thus it is possible to calculate each $\eta^{\left(r,k\right),i}_j$ matrix, use it, and discard it without needing to store it for future calculations.  Moreover, each $\frac{\partial f}{\partial w^{\left(k\right),i}_j}$ term has approximately the same number of entries as each $\eta^{\left(n,k\right),i}_j$ matrix, which means that even if it were necessary to store each $\eta^{\left(r,k\right),i}_j$, this would only be $\mathcal{O}\left(n\right)$ times as much information as a full gradient.  In other words, the amount of information involved would be $\mathcal{O}\left(nN\right)$, where $N$ is the total number of weights, rather than $\mathcal{O}\left(N^2\right)$, as would be expected in calculating a Hessian.

The key thing to note about these second derivatives is that they retain an outer product structure -- they are now tensor products (or the sums of tensor products) of matrices and vectors.  For example,

\begin{gather}
\frac{\partial^2 f}{\partial u^i_j \partial w^{\left(k\right),s}_t} = \left(\frac{\partial f}{\partial p^i} \times \eta^{\left(n,k\right),j}_s \times v^{\left(k-1\right),t}\right) + \left(a_{is} \times v^{\left(n\right),j} \times v^{\left(k-1\right),t}\right) \\
a_{is} = \frac{\partial^2 f}{\partial p^i \partial p^l} u^l_m \eta^{\left(n,k\right),m}_s
\end{gather}

It is important to note that this outer prodcut structure only exists \emph{for each sample}.  An average of a small number of samples (relative to the number of weights) will produce a low-rank structure; the rank will increase, generally, with the number of samples being considered.  As such, manipulating the entire Hessian may not be as computationally feasible; this will depend on how large the mini-batch size is relative to the number of weights.  When it comes to using the Hessian, the computational savings provided by this approach will be most salient when calculating scalar or vector quantities on a sample-by-sample basis and then taking a weighted sum of the results.  

In principle, we could calculate third derivatives, but the formulae would likely become unwieldy, and they may require memory usage significantly greater than that involved in storing gradient information.  If a use arose for third derivatives, calculating them would be conceivable, though.  Thus far, we have not included bias terms.  Including bias terms as trainable weights would increase the overall size of the gradient (by adding additional variables), but it would not change the overall structure.  Using the calculations provided in Appendix \ref{Low-Rank Derivations for Deep Networks}, it would not be difficult to produce the appropriate derivations.

This outer product structure also provides a way to calculate matrix-vector products with the Hessian.  Pearlmutter \cite{pearlmutter93tr}, for example, shows how these products can be calculated efficiently but does not distinguish between weights from different layers and thus does not observe or exploit the outer product structure.

\subsection{Convolutional Neural Networks}
\label{Convolutional Neural Networks}

Recurrent neural network derivatives exhibit an outer product structure similar to that of feedforward networks, but convolutional network derivatives do not; see Appendix \ref{Low-Rank Derivations for Deep Networks} for more details.  Convolutional layers do not destroy the outer product structure of subsequent feedforward or recurrent layers' derivatives, but they themselves do not have an outer product structure.  In fact, a convolutional layer is a different way of compressing weight information, as noted by Arora et al. \cite{arora18cp2}.  A convolutional layer is equivalent to a feedforward layer where the layer weights form a doubly block circulant matrix \cite{goodfellow16bk}.  The information contained in that (very large) matrix can be completely expressed by a much smaller matrix -- the convolution kernel, in fact.  This kernel enforces structure on weights and their derivatives, but the outer product structure on feedforward and recurrent layers only applies to the derivatives; convolutional layers tend to have fewer weights as a result.  Both types of structure, however, make it possible to represent derivative information in a compact fashion.  Since convolutional layers tend to have far fewer weights than feedforward layers, calculating higher-order derivatives with respect to these layers' weights should also be feasible.

\section{Discussion}

\subsection{Training Methods and Second-Order Information}
\label{Training Methods and Second-Order Information}

Deep learning (DL) provides a set of problems that can be tackled with gradient-based optimization methods, but it has a number of unique features and challenges.  Firstly, DL problems can be extremely large, and storing the Hessian, or even a full matrix approximation thereto, has not been considered feasible for such problems.  Secondly, DL problems are often highly nonconvex.  In practice, neural networks tend to have many saddle points and relatively few local minima that provide significantly suboptimal performance \cite{dauphin14cp}.  Thirdly, training deep networks via mini-batch sampling results in a stochastic optimization problem.  Even if the necessary expectations can be calculated (in an unbiased way), the variance associated with the batch sample calculations produces noise, and this noise can make it more difficult to perform the optimization.  Finally, deep networks consist of the composition of analytic functions whose forms are known.  As such, we can calculate derivative information analytically via back-propagation.

These special characteristics of DL have motivated researchers to develop training methods specifically designed to overcome the challenges with training a deep neural network.  One such approach is layer-wise pretraining \cite{bengio2007greedy}, where pretraining a neural network layer-by-layer encourages the weights to initialize close to a optimal minimum. Transfer learning \cite{yosinski2014transferable} works by a similar mechanism, relying on knowledge gained through previous tasks to encourage nice training on a novel task.  Outside of pretraining, a class of optimization algorithms have been specifically designed for training deep networks.  The Adam, Adagrad, and Adamax set of algorithms provide examples of using history-dependent learning rate adjustment \cite{kingma2014adam}.  Similarly, Nesterov momentum provides a method for leveraging history dependence in stochastic gradient descent \cite{sutskever2013importance}.  One could possibly argue that these methods implicitly leverage second order information via their history dependence, but the stochastic nature of mini-batching prevents this from becoming explicit.

Some researchers have sought to use second-order information explicitly to improve the training process.  Most of these methods have used an approximation to the Hessian.  For example, the L-BFGS method can estimate the Hessian (or its inverse) in a way that is feasible with respect to memory requirements; however, the noise associated with the sampling techniques can either overwhelm the estimation or require special modifications to the L-BFGS method to prevent it from diverging \cite{byrd16jsr}.  There have been two primary ways to deal with this: subsampling \cite{byrd16jsr,moritz16jsr} and mini-batch reuse \cite{schraudolph07jsr,mokhtari14jsr}.  Subsampling involves updating the Hessian approximation every $L$ iterations rather than every iteration, as would normally be done.  Mini-batch reuse consists of using the same mini-batch on subsequent iterations when calculating the difference in gradients between those two iterations.  These approximate second-order methods typically have a computational cost that is higher than, though on the same order of, gradient descent, and that cost can be further reduced by using a smaller mini-batch for the Hessian approximation calculations than for the gradient calculation \cite{byrd11jsr}.  There is also the question of bias: it is possible to produce unbiased low-rank Hessian approximations \cite{martens12cp}, but if the Hessian is indefinite, then quasi-Newton methods will prefer biased estimates -- ones that are positive definite.  Other work has foregone these kinds of Hessian approximations in favor of using finite differences \cite{martens2010deep}.

This outer product structure could be useful for helping with these kinds of calculations.  Storing the outer product components would make it possible to represent and manipulate the Hessian exactly.  For example, Dauphin et al. \cite{dauphin14cp} describe a way to use Newton's method and avoid saddle points when doing so.  However, they are forced to use an approximation: ``The exact implementation of this algorithm is intractable in a high dimensional problem, because it requires the exact computation of the Hessian. Instead we use an approach similar to Krylov subspace descent[.]'' \cite[p. 6]{dauphin14cp}.  The outer product structure could obviate the need for the approximation.  With the exact Hessian, it would also be possible to solve a problem like

\begin{gather}
\min_x x^T H x \\
\left\| x \right\| = 1
\end{gather}

\noindent to find the direction of maximum negative curvature and thereby escape from the saddle more quickly.  For the sample network given in Section \ref{Feedforward Network Derivative Calculations}, we can calculate this as follows:

\begin{gather}
\Delta u^i_j \equiv \omega^i_j \\
\Delta w^{\left(k\right),i}_j \equiv \phi^{\left(k\right),i}_j \\
x^T H x = \omega^i_j \frac{\partial^2 F}{\partial p^i \partial p^s} v^{\left(n\right),j} v^{\left(n\right),t} \omega^s_t \nonumber \\
+ 2 \sum_k \omega^i_j \left( \frac{\partial F}{\partial p^i} \eta^{\left(n,k\right),j}_s v^{\left(k-1\right),t} + \frac{\partial^2 F}{\partial p^i \partial p^l} v^{\left(n\right),j} u^l_m \eta^{\left(n,k\right),m}_s v^{\left(k-1\right),t} \right) \phi^{\left(k\right),s}_t \nonumber \\
+ \sum_k \phi^{\left(k\right),i}_j \frac{\partial^2 F}{\partial p^l \partial p^q} u^l_m \eta^{\left(n,k\right),m}_i v^{\left(k-1\right),j} u^q_a \eta^{\left(n,k\right),a}_s v^{\left(k-1\right),t} \phi^{\left(k\right),s}_t \nonumber \\
+ 2 \sum \limits_{k=2}^n \sum \limits_{r=1}^{k-1} \phi^{\left(k\right),i}_j \frac{\partial F}{\partial p^l} u^l_m \eta^{\left(n,k\right),m}_i \eta^{\left(k-1,r\right),j}_s v^{\left(r-1\right),t} \phi^{\left(r\right),s}_t 
\label{xHx eqn} \\
\left\| x \right\|^2 = \omega^i_j \omega^i_j + \sum_k \phi^{\left(k\right),i}_j \phi^{\left(k\right),i}_j
\end{gather}

The expression in Equation \ref{xHx eqn} looks complicated, but evaluating it is actually relatively straightforward -- it consists of a series of matrix multiplications where the matrices in question have a number of entries on the same order as the number of weights in a given layer.  For example, let us consider the last term in Equation \ref{xHx eqn}.  We can show that this Hessian component of this term breaks down into three separate components, each with no more than two indices:

\begin{gather}
\frac{\partial F}{\partial p^l} u^l_m \eta^{\left(n,k\right),m}_i \eta^{\left(k-1,r\right),j}_s v^{\left(r-1\right),t} = \left( \frac{\partial F}{\partial p^l} u^l_m \eta^{\left(n,k\right),m}_i \right) \times \eta^{\left(k-1,r\right),j}_s \times v^{\left(r-1\right),t} \nonumber \\
= \left(\cdot\right)_i \left(\cdot\right)^j_s \left(\cdot\right)^t
\end{gather}

Multiplying this with the appropriate $\phi$ terms is then simpler than would otherwise be expected:

\begin{gather}
\phi^{\left(k\right),i}_j \frac{\partial F}{\partial p^l} u^l_m \eta^{\left(n,k\right),m}_i \eta^{\left(k-1,r\right),j}_s v^{\left(r-1\right),t} \phi^{\left(r\right),s}_t = \phi^{\left(k\right),i}_j \left(\cdot\right)_i \left(\cdot\right)^j_s \left(\cdot\right)^t \phi^{\left(r\right),s}_t
\end{gather}

A similar decomposition can be performed for the other terms in Equation \ref{xHx eqn}.  Therefore, to perform this optimization, we need only calculate those components \emph{once}.  We can then re-evaluate $x^T Hx$ repeatedly for different values of $x$.  We would not see this kind of simplification if the Hessian expression had some irreducible $A^{jt}_{is}$ term -- the lack of such terms is precisely what we mean by saying that the Hessian has an outer product structure.

\subsection{Regularization and Robustness}

Higher-order derivatives could also be useful as a regularization tool.  Neural networks are known to be vulnerable to adversarial examples \cite{athalye18cp}.  The problem is that classifiers can have very large gradients, and as such, small pertubations in the inputs can cause large changes in the classifier.  A common defensive technique is to make it difficult for an attacker to evaluate those gradients; this is known as obfuscated gradients.  Athalye et al. showed that this approach may not be very robust, though \cite{athalye18cp}.  Some researchers have tried to regularize network training with respect to those gradients directly \cite{drucker92jsr,gu14cp,ross18cp}.  The gradient of this regularization then involves higher-order derivative terms, as Ross and Doshi-Velez note \cite{ross18cp}, which may make training infeasible.  Gu and Rigazio \cite{gu14cp}, for example, are forced to use approximations.  This may not be necessary, though.  Consider a regularization with respect to the gradients of the classifier output with respect to the inputs:

\begin{gather}
\frac{\partial f}{\partial v^{\left(0\right),i}} = \frac{\partial f}{\partial p^j} u^j_l \alpha^{\left(n,0\right),l}_i \equiv \zeta_i \\
\left\| \frac{\partial f}{\partial v^{\left(0\right),i}} \right\|^2 = \zeta_i \zeta_i \\
\frac{\partial}{\partial \left(\cdot\right)} \left( \left\| \frac{\partial f}{\partial v^{\left(0\right),i}} \right\|^2 \right) = 2 \zeta_i \frac{\partial \zeta_i}{\partial \left(\cdot\right)} \\
\frac{\partial \zeta_i}{\partial u^r_s} = \frac{\partial f}{\partial p^r} \alpha^{\left(n,0\right),s}_i + \frac{\partial^2 f}{\partial p^j \partial p^r} u^j_i v^{\left(n\right),s} \\
\frac{\partial \zeta_i}{\partial w^{\left(k\right),r}_s} = \frac{\partial f}{\partial p^j} u^j_l \eta^{\left(n,k\right),l}_r \alpha^{\left(k,0\right),s}_i + \frac{\partial^2 f}{\partial p^j \partial p^q} u^j_l \alpha^{\left(n,0\right),l}_i u^q_t \eta^{\left(n,k\right),t}_r v^{\left(k\right),s}
\end{gather}

Papernot et al. \cite{papernot16cp} claim that this kind of approach is only good when the perturbations are \emph{very} small and that the calculations are too expensive to be feasible.  We do want the network to be locally constant near our sample points, though \cite{goodfellow16bk}, and that cost may not be so prohibitive, now.  We can also extend this to a regularization on the curvature:

\begin{gather}
\frac{\partial^2 f}{\partial v^{\left(0\right),i} \partial v^{\left(0\right),r}} = \frac{\partial^2 f}{\partial p^j \partial p^q} u^j_l u^q_m \alpha^{\left(n,0\right),l}_i \alpha^{\left(n,0\right),m}_r \equiv \xi_{ir} \\
\left\| \frac{\partial^2 f}{\partial v^{\left(0\right),i} \partial v^{\left(0\right),r}} \right\|^2 = \xi_{ir} \xi_{ir} \\
\frac{\partial }{\partial \left(\cdot\right)} \left(\left\| \frac{\partial^2 f}{\partial v^{\left(0\right),i} \partial v^{\left(0\right),r}} \right\|^2 \right) = 2 \xi_{ir} \frac{\partial \xi_{ir}}{\partial \left(\cdot\right)} \\
\frac{\partial \xi_{ir}}{\partial u^s_t} = \frac{\partial^2 f}{\partial p^s \partial p^q} u^q_m \alpha^{\left(n,0\right),t}_i \alpha^{\left(n,0\right),m}_r + \frac{\partial^2 f}{\partial p^j \partial p^s} u^j_l \alpha^{\left(n,0\right),l}_i \alpha^{\left(n,0\right),t}_r \nonumber \\
+ \frac{\partial^3 f}{\partial p^j \partial p^q \partial p^s} u^j_l u^q_m \alpha^{\left(n,0\right),l}_i \alpha^{\left(n,0\right),m}_r v^{\left(n\right),t} \\
\frac{\partial \xi_{ir}}{\partial w^{\left(k\right),s}_t} = \frac{\partial^2 f}{\partial p^j \partial p^q} u^j_l u^q_m \left( \alpha^{\left(n,0\right),l}_i \eta^{\left(n,k\right),m}_s \alpha^{\left(k,0\right),t}_r + \eta^{\left(n,k\right),l}_s \alpha^{\left(k,0\right),t}_i \alpha^{\left(n,0\right),l}_r \right) \nonumber \\
+ \frac{\partial^3 f}{\partial p^j \partial p^q \partial p^a} u^j_l u^q_m u^a_b \alpha^{\left(n,0\right),l}_i \alpha^{\left(n,0\right),m}_r \eta^{\left(n,k\right),b}_s v^{\left(k\right),t}
\end{gather}

Both of these regularizations scale reasonably well as long as the number of classes is much smaller than the number of weights (which will generally be true).  For example, if there are 1000 weights and 10 categories, $\frac{\partial^3 f}{\partial \mathbf{p}^3}$ will have $10^3 = 1000$ entries -- the same number of entries as the gradients used for training.  The first-order regularization will require fewer multiplication operations and therefore be less computationally expensive, but there is not a significant difference in the amount of information that needs to be stored in order to calculate the gradients needed for training the weights.  Moreover, even within a given sample's calculation, the multiplications involved are highly parallelizable.

We can use this for training purposes, but we can also use this to provide estimated bounds on the size of perturbations needed for misclassification.  With the first-order approximation, we have

\begin{gather}
\Delta y^i \approx \frac{\partial y^i}{\partial x^j} \Delta x^j \Rightarrow \frac{\left\| \Delta y^i\right\|}{\left\|\frac{\partial y^i}{\partial x^j} \right\|} \leq \left\| \Delta x^j\right\| \\
\left\|\frac{\partial y^i}{\partial x^j} \right\| \leq \left\|\frac{\partial y^i}{\partial x^j} \right\|_F \Rightarrow \frac{\left\| \Delta y^i\right\|}{\left\|\frac{\partial y^i}{\partial x^j} \right\|_F} \leq \left\| \Delta x^j\right\|
\end{gather}

For the example network in Section \ref{Feedforward Network Derivative Calculations}, we would have

\begin{gather}
\frac{\partial y^i}{\partial v^{\left(0\right),j}} = \frac{\partial y^i}{\partial p^m} u^m_l \alpha^{\left(n,0\right),l}_j 
\end{gather}

A second-order approximation could function similarly, but there might be other ways to use that second-order information to provide an appropriate bound as well.  The point is that, as Athalye et al. note, we need specific, guaranteed robustness claims \cite{athalye18cp}.  Explicitly evaluating these gradients could be a way to do this.

\subsection{Compression and Generalizability}

We may be able to use this outer product structure to create networks that are more compressible and generalizable.  One common way to try and improve generalization is to regularize with respect to the weights (usually via some norm on the weights) \cite{goodfellow16bk}.  This is essentially a computable proxy for what we really want: what we really want is to avoid overfitting and ensure that small input changes produce small output changes.  Moody \cite{moody92cp} provides an approximate formula for relating training error to testing error.  This formula includes a measure of the effective number of parameters in the network, and Moody measures this using higher-order derivative information.  Reducing the magnitude of those derivatives essentially means reducing the effective number of parameters and bringing the testing and training errors closer together.  Directly regularizing this may now be possible.

Network pruning \cite{louizos17cp} and compression \cite{arora18cp2}, on the other hand, are both ways of reducing the actual number of parameters in a network to make the network more efficient (in some sense).  Compressibility essentially entails taking a low-rank approximation to the layer-wise weight matrices.  If we look at our gradient calculations, we can see that this does some interesting things.  Consider the feedforward network described in Section \ref{Feedforward Network Derivative Calculations}.  If we specify $u^i_j = a^i b_j$ and $w^{\left(k\right),i}_j = c^{\left(k\right),i} e^{\left(k\right)}_j$ (i.e., a rank-1, outer product \emph{weight} structure), then

\begin{gather}
\frac{\partial F}{\partial a^i} = \frac{\partial F}{\partial u^l_m} \frac{\partial u^l_m}{\partial a^i} = \frac{\partial F}{\partial p^i} v^{\left(n\right),j} b_j = \left(\cdot\right) \times \frac{\partial F}{\partial p^i}
\label{outer prod weights start}\\
\frac{\partial F}{\partial b_j} = \frac{\partial F}{\partial u^l_m} \frac{\partial u^l_m}{\partial b_j} = \frac{\partial F}{\partial p^l} v^{\left(n\right),j} a^l = \left(\cdot\right) \times v^{\left(n\right),j} \\
\frac{\partial F}{\partial c^{\left(k\right),i}} = \frac{\partial F}{\partial w^{\left(k\right),l}_m} \frac{\partial w^{\left(k\right),l}_m}{\partial c^{\left(k\right),i}} = \frac{\partial F}{\partial p^l} a^l b_m \eta^{\left(n,k\right),m}_i v^{\left(k-1\right),j} e^{\left(k\right)}_j = \left(\cdot\right) \times \frac{\partial F}{\partial p^l} a^l b_m \eta^{\left(n,k\right),m}_i \\
\frac{\partial F}{\partial e^{\left(k\right)}_j} = \frac{\partial F}{\partial w^{\left(k\right),l}_m} \frac{\partial w^{\left(k\right),l}_m}{\partial e^{\left(k\right)}_j} = \frac{\partial F}{\partial p^l} a^l b_m \eta^{\left(n,k\right),m}_i v^{\left(k-1\right),j} c^{\left(k\right),i} = \left(\cdot\right) \times v^{\left(k-1\right),j}
\label{outer prod weights end}
\end{gather}

\noindent where $\left(\cdot\right)$ indicates the presence of a scalar quantity.

Arora et al. \cite{arora18cp2} essentially say that these weight matrices become approximately low-rank near the end of training.  What if we were to use a low-rank or outer product weight matrix structure from the beginning of training?  The derivatives now become dense and much smaller (in terms of the number of entries), as can be seen in Equations \ref{outer prod weights start}-\ref{outer prod weights end}.  This could be a cost-efficient pre-training strategy.

Finally, this derivative structure may also suggest something about the loss function near critical points.  If there are more weights than samples (as is often the case), then the gradient and Hessian likely have non-trivial nullspaces at critical points (saddles as well as optima).  This would suggest that each critical point exists on a (locally) connected submanifold of critical points with (locally) constant loss function values.  Neural network training never actually converges exactly to a critical point \cite{goodfellow16bk}, but this still might tell us something about what to expect as we approach these critical points.

\section{Conclusions}

In this paper, we showed how neural network derivatives display outer product structure for a wide range of network architectures.  This structure does not seem to have been widely appreciated or used -- especially with regards to higher-order derivative calculations.  However, taking advantage of this structure could be helpful for improving training methods, increasing network robustness and generalizability, and compressing layer weights.

\section*{Acknowledgments}
\label{Acknowledgments}
This research was performed at the Pacific Northwest National Laboratory, a multi-program national laboratory operated by Battelle for the U.S. Department of Energy

\bibliography{bib}
\bibliographystyle{ieeetr}

\appendix

\section{Outer Product Derivations for Deep Networks}
\label{Low-Rank Derivations for Deep Networks}

\subsection{Convolutional and Recurrent Layers}
 
Convolutional and recurrent layers preserve the outer product derivative structure of the fully connected feedforward layers considered above, and we will show this in the following sections.  Because we are only considering a single layer of each, we calculate the derivatives of the layer outputs with respect to the layer inputs -- in a larger network, those derivatives will be necessary for calculating total derivatives via back-propagation.

\subsubsection{Convolutional Layer}

We can define a convolutional layer as

\begin{gather}
v^s_t = \mathcal{A}\left(\left(\tilde{x}^s_t\right)^l_k w^l_k\right) \\
\left(\tilde{x}^s_t\right)^l_k = x^{\sigma s + l -1}_{\tau t + k - 1}
\end{gather}

\noindent where $x^i_j$ is the layer input, $\sigma$ is the vertical stride, $\tau$ is the horizontal stride, $\mathcal{A}$ is the activation function, and $v^s_t$ is the layer output.  A convolutional structure can make the expressions somewhat complicated when expressed in index notation, but we can simplify matters by using the simplification $z^{sl}_{tk} = x^{\sigma s + l -1}_{\tau t + k - 1}$.  The layer definition is then

\begin{equation}
v^s_t = \mathcal{A}\left(z^{sl}_{tk} w^l_k\right)
\end{equation}

The derivatives of the convolutional layer are

\begin{gather}
\frac{\partial v^s_t}{\partial x^i_j} = \mathcal{A}'\left(z^{sl}_{tk} w^l_k\right) w^m_p \frac{\partial z^{sm}_{tq}}{\partial x^i_j} \\
\frac{\partial z^{sm}_{tq}}{\partial x^i_j} = \left\{ \begin{array}{cc}
1		&i = \sigma s + m -1 \text{ and } j = \tau t + p -1 \\
0		& \text{else} \end{array} \right. \\
\frac{\partial v^s_t}{\partial w^p_q} = \mathcal{A}'\left(z^{sl}_{tk} w^l_k\right) z^{sp}_{tq} \\
\frac{\partial^2 v^s_t}{\partial x^p_q \partial x^i_j} = \mathcal{A}''\left(z^{sl}_{tk} w^l_k\right) w^m_r \frac{\partial z^{sm}_{tr}}{\partial x^i_j}  w^a_b \frac{\partial z^{sa}_{tb}}{\partial x^p_q} \\
\frac{\partial^2 v^s_t}{\partial w^p_q \partial x^i_j} = \mathcal{A}'\left(z^{sl}_{tk} w^l_k\right) \frac{\partial z^{sp}_{tq}}{\partial x^i_j} + \mathcal{A}''\left(z^{sl}_{tk} w^l_k\right) z^{sp}_{tq} w^m_r \frac{\partial z^{sm}_{tr}}{\partial x^i_j} \\
\frac{\partial^2 v^s_t}{\partial w^p_q \partial w^a_b} = \mathcal{A}''\left(z^{sl}_{tk} w^l_k\right) z^{sp}_{tq} z^{sa}_{tb}
\end{gather}

\noindent with no summation over $s$ and $t$ in any of the expressions above.  Using the simplification with $z^{sl}_{tk}$ makes it easier to see the structure of these derivatives.

\begin{gather}
\frac{\partial v^s_t}{\partial w^p_q} = \mathcal{A}'\left(\left(\tilde{x}^s_t\right)^l_k w^l_k\right)\left(\tilde{x}^s_t\right)^p_q =  \mathcal{A}'\left(\left(\tilde{x}^s_t\right)^l_k w^l_k\right) x^{\sigma s + p -1}_{\tau t + q - 1}\\
\frac{\partial v^s_t}{\partial x^i_j} = \left\{ \begin{array}{cc}
\mathcal{A}'\left(\left(\tilde{x}^s_t\right)^l_k w^l_k\right) w^{i+1-\sigma s}_{j+1 - \tau t}		&i+1 > \sigma s, j+1 >\tau t \\
0																	&\text{else} \end{array} \right. \\
\frac{\partial^2 v^s_t}{\partial w^p_q \partial w^a_b} =  \mathcal{A}''\left(\left(\tilde{x}^s_t\right)^l_k w^l_k\right) x^{\sigma s + p -1}_{\tau t + q - 1} x^{\sigma s + a -1}_{\tau t + b - 1} \\
\frac{\partial^2 v^s_t}{\partial w^p_q \partial x^i_j} \nonumber \\
=  \left\{ \begin{array}{cc}
\mathcal{A}'\left(\left(\tilde{x}^s_t\right)^l_k w^l_k\right) +  \mathcal{A}''\left(\left(\tilde{x}^s_t\right)^l_k w^l_k\right) w^{i+1-\sigma s}_{j+1 - \tau t} x^{\sigma s + p -1}_{\tau t + q - 1}		&i+1 - \sigma s=p, j+1 -\tau t=q \\ \\
\mathcal{A}''\left(\left(\tilde{x}^s_t\right)^l_k w^l_k\right) w^{i+1-\sigma s}_{j+1 - \tau t} x^{\sigma s + p -1}_{\tau t + q - 1}	&\begin{array}{c} i+1 - \sigma s>0, i+1 - \sigma s \neq p, \\	j+1 -\tau t > 0, j+1 -\tau t \neq q \end{array} \\ \\
0																	&\text{else} \end{array} \right. \\
\frac{\partial^2 v^s_t}{\partial x^p_q \partial x^i_j} = \left\{ \begin{array}{cc}
\mathcal{A}''\left(\left(\tilde{x}^s_t\right)^l_k w^l_k\right) w^{i+1-\sigma s}_{j+1 - \tau t} w^{p+1-\sigma s}_{q+1 - \tau t}		&\begin{array}{c} p+1 > \sigma s,i+1 > \sigma s \\ q+1 > \tau t,j+1 >\tau t \end{array} \\ \\
0																	&\text{else} \end{array} \right.
\end{gather}

The conditional form of the expressions is more complicated, but it is also possible to see how the derivatives relate to $w^i_j$ and submatrices of $x^i_j$.

\subsubsection{Sequence of Convolutional Layers}

Index notation becomes unwieldy with multiple convolutional layers in sequence because of the nature of the convolution operation.  We describe a sequence of convolutional layers as

\begin{gather}
v\left(s,t;k\right) = \mathcal{A}\left( \sum_{l,j} w\left(l,j;k\right) v\left(\sigma_k s + l -1,\tau_k t + j - 1;k-1\right) \right) \equiv \mathcal{A}\left(s,t,l,j;k\right)
\end{gather}

\noindent where $k$ indicates the layer, $\sigma$ is the vertical stride, $\tau$ is the horizontal stride, $v$ is a layer output, and $w$ are layer weights.  The first derivatives are as follows:

\begin{gather}
\frac{\partial v\left(s,t;k\right)}{\partial w\left(l,j;p\right)} = \left\{ \begin{array}{cc}
0 																											&p>k \\
\mathcal{A}'\left(s,t;k\right) v\left(\sigma_k s + l - 1,\tau_k t + j - 1;k-1\right)		&l=k \\
\mathcal{A}'\left(s,t;k\right) w \left(q,r;k\right) \frac{\partial v\left(\sigma_k s + q - 1,\tau_k t + r - 1;k-1\right)}{\partial w\left(l,j;p\right)}			&l<k \end{array} \right. \\
\beta\left(s,t,q,r;k\right) = \mathcal{A}'\left(s,t;k\right) w \left(q,r;k\right) \\
\frac{\partial v\left(s_k,t_k;k\right)}{\partial w\left(q_p,r_p;p\right)} \nonumber \\
= \beta \left(s_k,t_k,q_k,r_k;k\right) \beta \left(s_{k-1},t_{k-1},q_{k-1},r_{k-1};k-1\right) \ldots \beta \left(s_{p+1},t_{p+1},q_{p+1},r_{p+1};p+1\right) \nonumber \\
\times \mathcal{A}'\left(s_p,t_p;p\right) v\left(\sigma_p s_p + q_p - 1,\tau_p t_p + r_p - 1;p-1\right)
\end{gather}

By defining the relationships between different indices, we can simplify this expression:

\begin{gather}
s_{k-1} = \sigma_{k-1} s_k + q_k - 1 \nonumber \\
= \left( \prod \limits_{m=1}^{n+1} \sigma_{k-m}\right) s_k + \sum \limits_{m=0}^{n-2} \left(\prod \limits_{a=m+2}^n \sigma_{k-a}\right) \left(q_{k-m} - 1\right) + \left(q_{k-n+1} - 1\right) \\
t_{k-1} = \tau_{k-1} t_k + r_k - 1 \nonumber \\
= \left( \prod \limits_{m=1}^{n+1} \tau_{k-m}\right) t_k + \sum \limits_{m=0}^{n-2} \left(\prod \limits_{a=m+2}^n \tau_{k-a}\right) \left(r_{k-m} - 1\right) + \left(r_{k-n+1} - 1\right) \\
\frac{\partial v\left(s_k,t_k;k\right)}{\partial w\left(q_p,r_p;p\right)} = \sum \limits_{m=p+1}^k \sum_{r_m,q_m} \left[ \left(\prod \limits_{a=p+1}^k \beta \left(s_a,t_a,q_a,r_a\right)\right) \right. \nonumber \\
\left. \times \mathcal{A}'\left(s_p,t_p;p\right) v\left(\sigma_p s_p + q_p - 1,\tau_p t_p + r_p - 1;p-1\right) \right]
\end{gather}

The convolutional layers, when combined, lose the outer product structure -- the indices being summed over extend through all of the terms.  This is directly connected to the way in which the receptive field of a sequence of convolutional layers changes with the number of layers (see Goodfellow et al. \cite{goodfellow16bk}).  To view it another way, feedforward networks preserve layer-wise outer product structure because each layer only interacts directly which the layers adjacent to it.  The convolutional structure, however, means that indices summed over in one layer show up in other layers as well.  Pooling layers are similar in how they compress information down.  Isolated pooling layers (e.g., in a feedforward network) should not unduly disrupt the outer product structure, just as an isolated convolutional layer would not be a problem, but connected pooling layers could be problematic in this regard.

\subsubsection{Recurrent Layer}

We can define our recurrent layer as

\begin{gather}
v^j_{\left(t\right)} = \mathcal{A} \left(w^j_i v^i_{\left(t-1\right)}\right) \\
v^j_{\left(0\right)} = x^j
\end{gather}

\noindent where $t$ indicates the number of times that the recursion has been looped through.  If we inspect this carefully, we can actually see that this is almost identical to the hidden layers of the feedforward network: they are identical if we stipulate that the weights of the feedforward network are identical at each layer (i.e. $w^{\left(k\right),i}_j = w^i_j \ \forall \ k$) and if we treat the recursive loops like layers.  This observation allows us to reuse some of our previous derivations.  Primarily, we will use the fact that

\begin{equation}
\frac{\partial \left(\cdot\right)}{\partial w^i_j} = \sum_k \frac{\partial \left(\cdot\right)}{\partial w^{\left(k\right),m}_p} \frac{\partial w^{\left(k\right),i}_j}{\partial w^i_j} = \sum_k \frac{\partial \left(\cdot\right)}{\partial w^{\left(k\right),i}_j}
\end{equation}

 The first-order derivatives are then

\begin{gather}
\frac{\partial v^m_{\left(t\right)}}{\partial w^s_r} = \sum_k \eta^{\left(t,k\right),m}_s v^r_{\left(t-1\right)} \\
\frac{\partial v^j_{\left(t\right)}}{\partial v^i_{\left(t-1\right)}} = \mathcal{A}'\left(w^j_i v^i_{\left(k-1\right)}\right) w^j_i \equiv \beta^{\left(t\right),j}_i \\
\frac{\partial v^{j_t}_{\left(t\right)}}{\partial v^{j_k}_{\left(k\right)}} = \beta^{\left(t\right),j_t}_{j_{t-1}} \frac{\partial v^{j_{t-1}}_{\left(t-1\right)}}{\partial v^{j_k}_{\left(k\right)}} \nonumber \\
= \prod \limits_{\tau=k+1}^t \beta^{\left(\tau\right),j_{\tau}}_{j_{\tau -1}} = \prod \limits_{\tau=k++21}^t \beta^{\left(\tau\right),j_{\tau}}_{j_{\tau -1}} \gamma^{\left(k+1\right),j_{k+1}}_s w^s_{j_k} = \eta^{\left(t,k+1\right),j_t}_s w^s_{j_k} \\
\frac{\partial v^m_{\left(t\right)}}{\partial x^i} = \eta^{\left(t,1\right),m}_s w^s_i
\end{gather}

If $\mathcal{A}$ is a ReLU, then the second derivatives are relatively simple

\begin{gather}
\frac{\partial v^m_{\left(t\right)}}{\partial w^i_j \partial w^q_r} = \sum_{k,p} \frac{\partial v^m_{\left(t\right)}}{\partial w^{\left(k\right),i}_j \partial w^{\left(p\right),q}_r} = 2 \sum \limits_{k=2}^t \sum \limits_{p=1}^{k-1} \eta^{\left(t,k\right),m}_i \eta^{\left(k-1,p\right),j}_q v^r_{\left(p-1\right)} \\
\frac{\partial v^m_{\left(t\right)}}{\partial x^i \partial x^l} = 0 \\
\frac{\partial v^m_{\left(t\right)}}{\partial w^q_r \partial x^i} = \eta^{\left(t,1\right),j}_s \delta^s_q \delta^r_i + \sum_k \frac{\partial \eta^{\left(t,1\right),m}_q}{\partial w^{\left(k\right),s}_t} w^s_i = \eta^{\left(t,1\right),j}_q \delta^r_i + \sum_k \eta^{\left(t,k\right),m}_q \eta^{\left(k-1,1\right),r}_s w^s_i
\end{gather}

If $\mathcal{A}$ is not a ReLU, then we would use the results in the next section to calculate the second derivatives.  Regardless of the exact form of $\mathcal{A}$, though, we retain the outer product structure as long as $\mathcal{A}$ is an entry-wise function of its arguments.

\subsubsection{Recurrent Network}

We can consider a recurrent neural network with a ReLU activation for the recurrent layer and softmax and categorical cross entropy used following that.  For this example, we an argument in the recurrent layer that differs somewhat from the previous section:

\begin{gather}
v^j_{\left(t\right)} = \mathcal{A} \left(w^j_i v^i_{\left(t-1\right)} + u^j_i x^i_{\left(t-1\right)}\right) \\
v^j_{\left(0\right)} = 0 \\
p^j = z^j_i v^i_{\left(n\right),i} \\
\hat{y}^j = \frac{ \exp \left(p^j\right)}{\sum \limits_l \exp\left(p^l\right)} \\
f = - y^l \ln \hat{y}^l \\
F = E\left[f\right]
\end{gather}

To aid in producing the derivations, we define intermediate quantities analogous to those used in the feedforward network:

\begin{gather}
\beta^j_{\left(t\right),k} \equiv \mathcal{A}'\left(w^j_i v^i_{\left(t-1\right)} + u^j_i x^i_{\left(t-1\right)}\right) w^j_k  = \gamma^j_{\left(t\right),m} w^m_k \\
\gamma^j_{\left(t\right),k} \equiv \mathcal{A}'\left(w^j_i v^i_{\left(t-1\right)} + u^j_i x^i_{\left(t-1\right)}\right) \delta^j_k \\
\alpha^{j_t}_{\left(t,r\right),j_{r-1}} \equiv \prod \limits_{s=r}^t \beta^{j_s}_{\left(s\right),j_{s-1}} ,  \ r \leq t \\
\alpha^j_{\left(t,t+1\right),m} \equiv \delta^j_m \\
\alpha^j_{\left(t,s\right),m} \equiv 0, \ s > t+1 \\
\alpha^j_{\left(t,t\right),m} = \beta^j_{\left(t\right),m} \\
\alpha^j_{\left(t,s\right),m} = \alpha^j_{\left(t,q\right),i} \alpha^i_{\left(q-1,s\right),m}, \ q-1 \leq t; s \leq q \\
\eta^j_{\left(t,r\right),k} \equiv \alpha^j_{\left(t,r\right),i} \gamma^i_{\left(r-1\right),k}
\end{gather}

In this formulation, we have weights $u^i_j$, $w^i_j$, and $z^i_j$ with input data $x^i_{\left(t\right)}$.  The first derivatives of the recurrent layer outputs are

\begin{gather}
\frac{\partial v^j_{\left(t\right)}}{\partial w^k_l} = \mathcal{A}'\left(w^j_i v^i_{\left(t-1\right)} + u^j_i x^i_{\left(t-1\right)}\right) \delta^j_k v^l_{\left(t-1\right)} + \mathcal{A}'\left(w^j_i v^i_{\left(t-1\right)} + u^j_i x^i_{\left(t-1\right)}\right) w^j_i \frac{\partial v^i_{\left(t-1\right)}}{\partial w^k_l} \nonumber \\
= \gamma^j_{\left(t\right),k} v^l_{\left(t-1\right)} + \beta^j_{\left(t\right),i} \frac{\partial v^i_{\left(t-1\right)}}{\partial w^k_l} \\
\frac{\partial v^{j_t}_{\left(t\right)}}{\partial w^k_l}  = \gamma^{j_t}_{\left(t\right),k} v^l_{\left(t-1\right)} + \sum \limits_{r = 2}^t \left( \prod \limits_{s=r}^t \beta^{j_s}_{\left(s\right),j_{s-1}} \right) \gamma^{j_{r-1}}_{\left(r-1\right),k} v^l_{\left(r-2\right)}  \nonumber \\
= \gamma^{j_t}_{\left(t\right),k} v^l_{\left(t-1\right)} + \sum \limits_{r = 2}^t \alpha^{j_r}_{\left(t,r\right),j_{r-1}} \gamma^{j_{r-1}}_{\left(r-1\right),k} v^l_{\left(r-2\right)} \nonumber \\
=  \sum \limits_{r = 2}^{t+1} \alpha^{j_r}_{\left(t,r\right),j_{r-1}} \gamma^{j_{r-1}}_{\left(r-1\right),k} v^l_{\left(r-2\right)} \nonumber \\
= \sum \limits_{r = 2}^{t+1} \eta^{j_r}_{\left(t,r\right),k} v^l_{\left(r-2\right)} \\
\frac{\partial v^j_{\left(t\right)}}{\partial u^k_l} = \mathcal{A}'\left(w^j_i v^i_{\left(t-1\right)} + u^j_i x^i_{\left(t-1\right)}\right) \delta^j_k x^l_{\left(t-1\right)} + \mathcal{A}'\left(w^j_i v^i_{\left(t-1\right)} + u^j_i x^i_{\left(t-1\right)}\right) w^j_i \frac{\partial v^i_{\left(t-1\right)}}{\partial u^k_l} \nonumber \\
= \gamma^j_{\left(t\right),k} x^l_{\left(t-1\right)} + \beta^j_{\left(t\right),i} \frac{\partial v^i_{\left(t-1\right)}}{\partial u^k_l} \\
\frac{\partial v^{j_t}_{\left(t\right)}}{\partial u^k_l} = \gamma^{j_t}_{\left(t\right),k} x^l_{\left(t-1\right)} + \sum \limits_{r = 2}^t \left( \prod \limits_{s=r}^t \beta^{j_s}_{\left(s\right),j_{s-1}} \right) \gamma^{j_{r-1}}_{\left(r-1\right),k} x^l_{\left(r-2\right)} \nonumber \\
= \gamma^{j_t}_{\left(t\right),k} x^l_{\left(t-1\right)} + \sum \limits_{r = 2}^t \alpha^{j_r}_{\left(t,r\right),j_{r-1}} \gamma^{j_{r-1}}_{\left(r-1\right),k} x^l_{\left(r-2\right)} \nonumber \\
=  \sum \limits_{r = 2}^{t+1} \alpha^{j_r}_{\left(t,r\right),j_{r-1}} \gamma^{j_{r-1}}_{\left(r-1\right),k} x^l_{\left(r-2\right)} \nonumber \\ 
= \sum \limits_{r = 2}^{t+1} \eta^{j_r}_{\left(t,r\right),k} x^l_{\left(r-2\right)}
\end{gather}

The network derivatives are then

\begin{gather}
\frac{\partial f}{\partial z^i_j} = \frac{\partial f}{\partial p^i} v^j_{\left(n\right)}\\
\frac{\partial f}{\partial w^i_j} = \frac{\partial f}{\partial p^k} z^k_l \frac{\partial v^l_{\left(n\right)}}{\partial w^i_j} = \sum \limits_{r = 2}^{n+1} \frac{\partial f}{\partial p^k} z^k_l \eta^l_{\left(n,r\right),i} v^j_{\left(r-2\right)} \\
\frac{\partial f}{\partial u^i_j} = \frac{\partial f}{\partial p^k} z^k_l \frac{\partial v^l_{\left(n\right)}}{\partial u^i_j} = \sum \limits_{r = 2}^{n+1} \frac{\partial f}{\partial p^k} z^k_l \eta^l_{\left(n,r\right),i} x^j_{\left(r-2\right)}
\end{gather}

Note that for ReLU activation functions, $\gamma^i_{\left(t\right),k}$ is a constant, and $\beta^j_{\left(t\right),k}$ is a $w^j_k$ multiplied by a constant.  Therefore the second derivatives in this case are

\begin{gather}
\frac{\partial^2 f}{\partial z^i_j \partial z^q_s} = \frac{\partial^2 f}{\partial p^i \partial p^q} v^j_{\left(n\right)} v^s_{\left(n\right)} \\
\frac{\partial^2 f}{\partial z^i_j \partial u^q_s} = \frac{\partial f}{\partial p^i} \frac{\partial v^j_{\left(n\right)}}{\partial u^q_s} + \frac{\partial^2 f}{\partial p^i \partial p^k} z^k_m \frac{\partial v^m_{\left(n\right)}}{\partial u^q_s} v^j_{\left(n\right)} \nonumber \\
= \sum \limits^{n+1}_{r=2} \left( \frac{\partial f}{\partial p^i} \eta^j_{\left(n,r\right),q} x^s_{\left(r-2\right)} + \frac{\partial^2 f}{\partial p^i \partial p^k} z^k_m \eta^m_{\left(n,r\right),q} x^s_{\left(r-2\right)} v^j_{\left(n\right)} \right) \\
\frac{\partial^2 f}{\partial z^i_j \partial w^q_s} = \frac{\partial f}{\partial p^i} \frac{\partial v^j_{\left(n\right)}}{\partial w^q_s} + \frac{\partial^2 f}{\partial p^i \partial p^k} z^k_m \frac{\partial v^m_{\left(n\right)}}{\partial w^q_s} v^j_{\left(n\right)} \nonumber \\
= \sum \limits^{n+1}_{r=2} \left( \frac{\partial f}{\partial p^i} \eta^j_{\left(n,r\right),q} v^s_{\left(r-2\right)} + \frac{\partial^2 f}{\partial p^i \partial p^k} z^k_m \eta^m_{\left(n,r\right),q} v^s_{\left(r-2\right)} v^j_{\left(n\right)} \right) 
\end{gather}

\begin{gather}
\frac{\partial \beta^j_{\left(t\right),k}}{\partial u^q_s} = 0 \Rightarrow \frac{\partial \alpha^j_{\left(t,r\right),k}}{\partial u^q_s} = \frac{\partial \eta^j_{\left(t,r\right),k}}{\partial u^q_s} = 0 \\
\frac{\partial^2 f}{\partial u^i_j \partial u^q_s} = \sum \limits_{r=2}^{n+1} \sum \limits_{c=2}^{n+1} \frac{\partial^2 f}{\partial p^k \partial p^a} z^k_l \eta^l_{\left(n,r\right),i} x^j_{\left(r-2\right)} z^a_b \eta^b_{\left(n,c\right),q} x^s_{\left(c-2\right)} \\
\frac{\partial^2 f}{\partial w^i_j \partial u^q_s} = \sum \limits_{r = 2}^{n+1} \left[ \sum \limits_{c=2}^{n+1} \left(\frac{\partial^2 f}{\partial p^k \partial p^a} z^k_l \eta^l_{\left(n,r\right),i} v^j_{\left(r-2\right)} z^a_b \eta^b_{\left(n,c\right),q} x^s_{\left(c-2\right)} \right) \right. \nonumber \\
\left. + \sum \limits_{c=2}^{r-1} \left( \frac{\partial f}{\partial p^k} z^k_l \eta^l_{\left(n,r\right),i} \eta^j_{\left(r-2,c\right),q} x^s_{\left(c-2\right)} \right) \right]
\end{gather}

\begin{gather}
\frac{\partial \beta^j_{\left(t\right),k}}{\partial w^q_s} = \mathcal{A}'\left(w^j_i v^i_{\left(t-1\right)} + u^j_i x^i_{\left(t-1\right)}\right) \delta^j_q \delta^s_k = \gamma^j_{\left(t\right),q} \delta^s_k \\
\alpha^j_{\left(t,r\right),i} = \alpha^j_{\left(t,c\right),l} \alpha^l_{\left(c-1,c-1\right),m} \alpha^m_{\left(c-2,r\right),i} = \alpha^j_{\left(t,c\right),l} \beta^l_{\left(c-1\right),m} \alpha^m_{\left(c-2,r\right),i}, \ r+1\leq c \leq t+1 \\
\frac{\partial \alpha^j_{\left(t,r\right),i}}{\partial w^q_s} = \sum \limits_{c=r+1}^{t+1} \alpha^j_{\left(t,c\right),l} \gamma^l_{\left(c-1\right),q} \delta^s_k \alpha^k_{\left(c-2,r\right),i} = \sum \limits_{c=r+1}^{t+1} \alpha^j_{\left(t,c\right),l} \gamma^l_{\left(c-1\right),q} \alpha^s_{\left(c-2,r\right),i}  \\
\frac{\partial \eta^j_{\left(t,r\right),i}}{\partial w^q_s} = \frac{\partial \alpha^j_{\left(t,r\right),m}}{\partial w^q_s} \gamma^m_{\left(r-1\right),i} = \sum \limits_{c=r+1}^{t+1} \eta^j_{\left(t,c\right),q} \eta^s_{\left(c-2,r\right),i} \\
\frac{\partial^2 f}{\partial w^i_j \partial w^q_s} = \sum \limits_{r=2}^{n+1} \left[ \sum \limits_{c=2}^{n+1} \left(\frac{\partial^2 f}{\partial p^k \partial p^a} z^k_l \eta^l_{\left(n,r\right),i} v^j_{\left(r-2\right)} z^a_b \eta^b_{\left(n,c\right),q} v^s_{\left(c-2\right)} \right) \right. \nonumber \\
\left. + \sum \limits_{c=r+1}^{n+1} \left(\frac{\partial f}{\partial p^k} z^k_l \eta^l_{\left(n,c\right),q} \eta^s_{\left(c-2,r\right),i} v^j_{\left(r-2\right)} \right) + \sum \limits_{c=2}^{r-1} \left(\frac{\partial f}{\partial p^k} z^k_l \eta^l_{\left(n,r\right),i} \eta^j_{\left(r-2,c\right),q} v^s_{\left(c-2\right)} \right) \right]
\end{gather}

\subsection{Deep Network with General Activation Functions}

For a deep feedforward network with general entry-wise activation functions, the first derivatives are all identical to the derivations given in Section \ref{Feedforward Network Derivative Calculations} save that $\mathcal{A}'$ and $\mathcal{A}''$ will be different.  Before calculating second-order derivatives, though, we do some preliminary calculations that were not necessary before because $\mathcal{A}'' = 0$ for ReLUs.  First, we calculate the derivatives of $\gamma^{\left(l\right),m}_r$:

\begin{gather}
\frac{\partial \gamma^{\left(l\right),m}_r}{\partial w^{\left(q\right),s}_t} = \left\{ \begin{array}{cc}
0		&l < q \\
\delta^m_r \mathcal{A}''\left(w^{\left(l\right),m}_j v^{\left(l-1\right),j}\right) \delta^m_s v^{\left(l-1\right),t}		&l=q \\
\delta^m_r \mathcal{A}''\left(w^{\left(l\right),m}_j v^{\left(l-1\right),j}\right) w^{\left(l\right),m}_a \eta^{\left(l-1,q\right),a}_s v^{\left(q-1\right),t}	&l>q
\end{array} \right. \\
\lambda^{\left(l\right),m}_{rs} \equiv \delta^m_r \delta^m_s \mathcal{A}''\left(w^{\left(l\right),m}_j v^{\left(l-1\right),j}\right) \\
\frac{\partial \gamma^{\left(l\right),m}_r}{\partial w^{\left(q\right),s}_t} = \left\{ \begin{array}{cc}
0		&l < q \\
\lambda^{\left(l\right),m}_{rs} v^{\left(l-1\right),t}		&l=q \\
\lambda^{\left(l\right),m}_{rp} w^{\left(l\right),p}_a \eta^{\left(l-1,q\right),a}_s v^{\left(q-1\right),t}	&l>q
\end{array} \right.
\end{gather}

\noindent where there is no summation over $m$ in any of these equations.  Next, we calculate the derivatives of $\alpha^{\left(k,l\right),j}_m$:

\begin{gather}
\alpha^{\left(k,l\right),j}_m = \alpha^{\left(k,r\right),j}_b \alpha^{\left(r,r-1\right),b}_a \alpha^{\left(r-1,l\right),a}_m \\
\alpha^{\left(r,r-1\right),b}_a = \beta^{\left(r\right),b}_a = \gamma^{\left(r\right),b}_s w^{\left(r\right),s}_a \\
\frac{\partial \beta^{\left(r\right),b}_a}{\partial w^{\left(q\right),s}_t} = \left\{ \begin{array}{cc}
0		& r<q \\
\lambda^{\left(r\right),b}_{ps} w^{\left(r\right),p}_a v^{\left(r-1\right),t} + \gamma^{\left(r\right),b}_s \delta^t_a		& r=q \\
\lambda^{\left(r\right),b}_{dp} w^{\left(r\right),d}_a w^{\left(r\right),p}_m \eta^{\left(r-1,q\right),m}_s v^{\left(q-1\right),t}		& r>q
\end{array} \right. \\
\frac{\partial \alpha^{\left(k,l\right),j}_m}{\partial w^{\left(q\right),s}_t} = \sum \limits_{r=l+1}^k \alpha^{\left(k,r\right),j}_b \frac{\partial \beta^{\left(r\right),b}_a}{\partial w^{\left(q\right),s}_t} \alpha^{\left(r-1,l\right),a}_m \nonumber \\ \nonumber \\
= \left\{ \begin{array}{cc}
0				& k<q \\ \\
\begin{array}{c}\sum \limits_{r=q+1}^k \alpha^{\left(k,r\right),j}_b \lambda^{\left(r\right),b}_{dp} w^{\left(r\right),d}_a w^{\left(r\right),p}_c \eta^{\left(r-1,q\right),c}_s v^{\left(q-1\right),t} \alpha^{\left(r-1,l\right),a}_m \\
+ \alpha^{\left(k,q\right),j}_b \left(\lambda^{\left(q\right),b}_{ps} w^{\left(q\right),p}_a v^{\left(q-1\right),t} + \gamma^{\left(q\right),b}_s \delta^t_a\right) \alpha^{\left(q-1,l\right),a}_m \end{array} 		&l<q\leq k \\ \\
\sum \limits_{r=l+1}^k \alpha^{\left(k,r\right),j}_b \lambda^{\left(r\right),b}_{dp} w^{\left(r\right),d}_a w^{\left(r\right),p}_c \eta^{\left(r-1,q\right),c}_s v^{\left(q-1\right),t} \alpha^{\left(r-1,l\right),a}_m		&q \leq l
\end{array} \right.
\end{gather}

Thirdly, we calculate the derivatives of $\eta^{\left(n,k\right),j}_l$:

\begin{gather}
\frac{\partial \eta^{\left(n,k\right),j}_l}{\partial w^{\left(q\right),s}_t} = \frac{\partial \alpha^{\left(n,k\right),j}_m}{\partial w^{\left(q\right),s}_t} \gamma^{\left(k\right),m}_l + \alpha^{\left(n,k\right),j}_m \frac{\partial \gamma^{\left(k\right),m}_l}{\partial w^{\left(q\right),s}_t} \\
\frac{\partial \eta^{\left(n,k\right),j}_l}{\partial w^{\left(q\right),s}_t} \nonumber \\
= \left\{ \begin{array}{cc}
\begin{array}{c}\sum \limits_{r=q+1}^n \alpha^{\left(n,r\right),j}_b \lambda^{\left(r\right),b}_{dp} w^{\left(r\right),d}_a w^{\left(r\right),p}_c \eta^{\left(r-1,q\right),c}_s v^{\left(q-1\right),t} \alpha^{\left(r-1,k\right),a}_m \gamma^{\left(k\right),m}_l \\
+ \alpha^{\left(n,q\right),j}_b \left(\lambda^{\left(q\right),b}_{ps} w^{\left(q\right),p}_a v^{\left(q-1\right),t} + \gamma^{\left(q\right),b}_s \delta^t_a\right) \alpha^{\left(q-1,k\right),a}_m \gamma^{\left(k\right),m}_l \end{array}		&q>k \\ \\
\begin{array}{c} \sum \limits_{r=k+1}^n \alpha^{\left(n,r\right),j}_b \lambda^{\left(r\right),b}_{dp} w^{\left(r\right),d}_a w^{\left(r\right),p}_c \eta^{\left(r-1,k\right),c}_s v^{\left(k-1\right),t} \alpha^{\left(r-1,k\right),a}_m \gamma^{\left(k\right),m}_l \\
+ \alpha^{\left(n,k\right),j}_m \lambda^{\left(k\right),m}_{ls} v^{\left(k-1\right),t} \end{array}			&q=k \\ \\
\begin{array}{c} \sum \limits_{r=k+1}^n \alpha^{\left(n,r\right),j}_b \lambda^{\left(r\right),b}_{dp} w^{\left(r\right),d}_a w^{\left(r\right),p}_c \eta^{\left(r-1,k\right),c}_s v^{\left(k-1\right),t} \alpha^{\left(r-1,k\right),a}_m \gamma^{\left(k\right),m}_l \\
+ \alpha^{\left(n,k\right),j}_m \lambda^{\left(k\right),m}_{lp} w^{\left(k\right),p}_a \eta^{\left(k-1,q\right),a}_s v^{\left(q-1\right),t}	\end{array}	&q<k
\end{array} \right. \\
= \left\{ \begin{array}{cc}
\begin{array}{c} \sum \limits_{r=q+1}^n \alpha^{\left(n,r\right),j}_b \lambda^{\left(r\right),b}_{dp} w^{\left(r\right),d}_a w^{\left(r\right),p}_c \eta^{\left(r-1,q\right),c}_s v^{\left(q-1\right),t} \eta^{\left(r-1,k\right),a}_l \\
+ \alpha^{\left(n,q\right),j}_b \lambda^{\left(q\right),b}_{ps} w^{\left(r\right),p}_a v^{\left(q-1\right),t} \eta^{\left(q-1,k\right),a}_l + \eta^{\left(n,q\right),j}_s \eta^{\left(q-1,k\right),t}_l \end{array} 		&q>k \\ \\
\begin{array}{c} \sum \limits_{r=k+1}^n \alpha^{\left(n,r\right),j}_b \lambda^{\left(r\right),b}_{dp} w^{\left(r\right),d}_a w^{\left(r\right),p}_c \eta^{\left(r-1,k\right),c}_s v^{\left(k-1\right),t} \eta^{\left(r-1,k\right),a}_l \\
+ \alpha^{\left(n,k\right),j}_m \lambda^{\left(k\right),m}_{ls} v^{\left(k-1\right),t} \end{array}			&q=k \\
\begin{array}{c} \sum \limits_{r=k+1}^n \alpha^{\left(n,r\right),j}_b \lambda^{\left(r\right),b}_{dp} w^{\left(r\right),d}_a w^{\left(r\right),p}_c \eta^{\left(r-1,k\right),c}_s v^{\left(k-1\right),t} \eta^{\left(r-1,k\right),a}_l \\
+ \alpha^{\left(n,k\right),j}_m \lambda^{\left(k\right),m}_{lp} w^{\left(k\right),p}_a \eta^{\left(k-1,q\right),a}_s v^{\left(q-1\right),t} \end{array}		&q<k
\end{array} \right.
\end{gather}

\begin{gather}
\frac{\partial^2 v^{\left(n\right),j}}{\partial w^{\left(k\right),l}_i \partial w^{\left(q\right),s}_t} = \left\{ \begin{array}{cc}
\frac{\partial \eta^{\left(n,k\right),j}_l}{\partial w^{\left(q\right),s}_t} v^{\left(k-1\right),i}		&q\geq k \\
\frac{\partial \eta^{\left(n,k\right),j}_l}{\partial w^{\left(q\right),s}_t} v^{\left(k-1\right),i} + \eta^{\left(n,k\right),j}_l \frac{\partial v^{\left(k-1\right),i}}{\partial w^{\left(q\right),s}_t}		& q<k
\end{array} \right. \\ \nonumber \\
= \left\{ \begin{array}{cc}
\begin{array}{c} \sum \limits_{r=q+1}^n \alpha^{\left(n,r\right),j}_b \lambda^{\left(r\right),b}_{dp} w^{\left(r\right),d}_a w^{\left(r\right),p}_c \eta^{\left(r-1,q\right),c}_s v^{\left(q-1\right),t} \eta^{\left(r-1,k\right),a}_l v^{\left(k-1\right),i}			\\
+ \alpha^{\left(n,q\right),j}_b \lambda^{\left(q\right),b}_{ps} w^{\left(q\right),p}_a v^{\left(q-1\right),t} \eta^{\left(q-1,k\right),a}_l v^{\left(k-1\right),i}	 + \eta^{\left(n,q\right),j}_s \eta^{\left(q-1,k\right),t}_l v^{\left(k-1\right),i}	\end{array} 		&q>k \\ \\
\begin{array}{c} \sum \limits_{r=k+1}^n \alpha^{\left(n,r\right),j}_b \lambda^{\left(r\right),b}_{dp} w^{\left(r\right),d}_a w^{\left(r\right),p}_c \eta^{\left(r-1,k\right),c}_s v^{\left(k-1\right),t} \eta^{\left(r-1,k\right),a}_l v^{\left(k-1\right),i} 		\\
+ \alpha^{\left(n,k\right),j}_m \lambda^{\left(k\right),m}_{ls} v^{\left(k-1\right),t} v^{\left(k-1\right),i} \end{array}				&q=k \\ \\
\begin{array}{c} \sum \limits_{r=k+1}^n \alpha^{\left(n,r\right),j}_b \lambda^{\left(r\right),b}_{dp} w^{\left(r\right),d}_a w^{\left(r\right),p}_c \eta^{\left(r-1,k\right),c}_s v^{\left(k-1\right),t} \eta^{\left(r-1,k\right),a}_l v^{\left(k-1\right),i}	\\
+ \alpha^{\left(n,k\right),j}_m \lambda^{\left(k\right),m}_{lp} w^{\left(k\right),p}_a \eta^{\left(k-1,q\right),a}_s v^{\left(q-1\right),t}v^{\left(k-1\right),i}	+ \eta^{\left(n,k\right),j}_l \eta^{\left(k-1,q\right),i}_s v^{\left(q-1\right),t} \end{array}		&q<k
\end{array} \right. 
\end{gather}

The second-order derivatives of the objective function are then

\begin{gather}
\frac{\partial^2 f}{\partial u^i_j \partial u^s_t} = \frac{\partial^2 f}{\partial p^k \partial p^l} \frac{\partial p^k}{\partial u^i_j} \frac{\partial p^l}{\partial u^s_t} = \frac{\partial^2 f}{\partial p^i \partial p^s} v^{\left(n\right),j} v^{\left(n\right),t} \\
\frac{\partial^2 f}{\partial u^i_j \partial w^{\left(k\right),s}_t} = \frac{\partial f}{\partial p^i} \frac{\partial v^{\left(n\right),j}}{\partial w^{\left(k\right),s}_t} + \frac{\partial^2 f}{\partial p^i \partial p^l} v^{\left(n\right),j} \frac{\partial p^l}{\partial v^{\left(n\right),m}} \frac{\partial v^{\left(n\right),m}}{\partial w^{\left(k\right),s}_t} \nonumber \\
= \frac{\partial f}{\partial p^i} \eta^{\left(n,k\right),j}_s v^{\left(k-1\right),t} + \frac{\partial^2 f}{\partial p^i \partial p^l} v^{\left(n\right),j} u^l_m \eta^{\left(n,k\right),m}_s v^{\left(k-1\right),t} \\
\frac{\partial^2 f}{\partial w^{\left(k\right),l}_i \partial w^{\left(q\right),s}_t} = \frac{\partial^2 f}{\partial p^a \partial p^b} \frac{\partial p^a}{\partial v^{\left(n\right),j}} \frac{\partial v^{\left(n\right),j}}{\partial w^{\left(k\right),l}_i} \frac{\partial p^b}{\partial v^{\left(n\right),c}} \frac{\partial v^{\left(n\right),c}}{\partial w^{\left(q\right),s}_t} + \frac{\partial f}{\partial p^m} \frac{\partial p^m}{\partial v^{\left(n\right),j}} \frac{\partial^2 v^{\left(n\right),j}}{\partial w^{\left(k\right),l}_i \partial w^{\left(q\right),s}_t} \nonumber \\
= \frac{\partial^2 f}{\partial p^a \partial p^b} u^a_j \eta^{\left(n,k\right),j}_l v^{\left(k-1\right),i} u^b_c \eta^{\left(n,q\right),c}_s v^{\left(q-1\right),t} + \frac{\partial f}{\partial p^a} u^a_j \frac{\partial^2 v^{\left(n\right),j}}{\partial w^{\left(k\right),l}_i \partial w^{\left(q\right),s}_t}
\end{gather}

\end{document}